\documentclass[preprint,12pt]{elsarticle}

\usepackage{listings}
\usepackage{color}

\definecolor{dkgreen}{rgb}{0,0.6,0}
\definecolor{gray}{rgb}{0.5,0.5,0.5}
\definecolor{mauve}{rgb}{0.58,0,0.82}

\usepackage{tikz}
\usetikzlibrary{backgrounds}
\usepackage{graphicx}
\usepackage{bm}
\usepackage{amsmath}
\usepackage{setspace}
\usepackage{hyperref}
\usepackage{subfigure}
\usepackage{multirow}
\hypersetup{
    colorlinks=true,       
}
\usepackage{amssymb}

\usepackage{lineno}
\usepackage{diagbox}

\journal{Journal Name}

\begin{document}

\begin{frontmatter}


\title{Deep Hidden Physics Models: Deep Learning of Nonlinear Partial Differential Equations}




\author{Maziar Raissi}
\address{Division of Applied Mathematics, Brown University,\\ Providence, RI, 02912, USA}

\begin{abstract}
A long-standing problem at the interface of artificial intelligence and applied mathematics is to devise an algorithm capable of achieving human level or even superhuman proficiency in transforming observed data into predictive mathematical models of the physical world. In the current era of abundance of data and advanced machine learning capabilities, the natural question arises: How can we automatically uncover the underlying laws of physics from high-dimensional data generated from experiments? In this work, we put forth a deep learning approach for discovering nonlinear partial differential equations from scattered and potentially noisy observations in space and time. Specifically, we approximate the unknown solution as well as the nonlinear dynamics by two deep neural networks. The first network acts as a prior on the unknown solution and essentially enables us to avoid numerical differentiations which are inherently ill-conditioned and unstable. The second network represents the nonlinear dynamics and helps us distill the mechanisms that govern the evolution of a given spatiotemporal data-set. We test the effectiveness of our approach for several benchmark problems spanning a number of scientific domains and demonstrate how the proposed framework can help us accurately learn the underlying dynamics and forecast future states of the system. In particular, we study the Burgers', Korteweg-de Vries (KdV), Kuramoto-Sivashinsky, nonlinear Schr\"{o}dinger, and Navier-Stokes equations.
\end{abstract}


\end{frontmatter}


\section{Introduction}
Recent advances in machine learning in addition to new data recordings and sensor technologies have the potential to revolutionize our understanding of the physical world in modern application areas such as neuroscience, epidemiology, finance, and dynamic network analysis where first-principles derivations may be intractable \cite{Rudye1602614}. In particular, many concepts from statistical learning can be integrated with classical methods in applied mathematics to help us discover sufficiently sophisticated and accurate mathematical models of complex dynamical systems directly from data. This integration of nonlinear dynamics and machine learning opens the door for principled methods for model construction, predictive modeling, nonlinear control, and reinforcement learning strategies. The literature on data-driven discovery of dynamical systems \cite{crutchfield1987equations} is vast and encompasses equation-free modeling \cite{kevrekidis2003equation}, artificial neural networks \cite{raissi2018multistep,gonzalez1998identification, anderson1996comparison, rico1992discrete}, nonlinear regression \cite{voss1999amplitude}, empirical dynamic modeling \cite{sugihara2012detecting, ye2015equation}, modeling emergent behavior \cite{roberts2014model}, automated inference of dynamics \cite{schmidt2011automated, daniels2015automated, daniels2015efficient}, normal form identification in climate \cite{majda2009normal}, nonlinear Laplacian spectral analysis \cite{giannakis2012nonlinear}, modeling emergent behavior \cite{roberts2014model}, Koopman analysis \cite{mezic2005spectral, budivsic2012applied, mezic2013analysis, brunton2017chaos}, automated inference of dynamics \cite{schmidt2011automated, daniels2015automated, daniels2015efficient}, and symbolic regression \cite{bongard2007automated, schmidt2009distilling}. More recently, sparsity \cite{tibshirani1996regression} has been used to determine the governing dynamical system \cite{brunton2016discovering, mangan2016inferring, wang2011predicting, schaeffer2013sparse, ozolicnvs2013compressed, mackey2014compressive, brunton2014compressive, proctor2014exploiting, bai2014low, tran2016exact}.\\

Less well studied is how to discover the underlying physical laws expressed by partial differential equations from scattered data collected in space and time. Inspired by recent developments in {\em physics-informed deep learning} \cite{raissi2017physics_I,raissi2017physics_II}, we construct structured nonlinear regression models that can uncover the  dynamic dependencies in a given set of spatio-temporal dataset, and return a closed form model that can be subsequently used to forecast future states. In contrast to recent approaches to systems identification \cite{brunton2016discovering,Rudye1602614}, here we do not need to have direct access or approximations to temporal or spatial derivatives. Moreover, we are using a richer class of function approximators to represent the nonlinear dynamics and consequently we do not have to commit to a particular family of basis functions. Specifically, we consider nonlinear partial differential equations of the general form
\begin{eqnarray}\label{eq:PDE}
&&u_t = \mathcal{N}(t,x,u,u_x,u_{xx},\ldots),
\end{eqnarray}
where $\mathcal{N}$ is a nonlinear function of time $t$, space $x$, solution $u$ and its derivatives.\footnote{ The solution $u = (u_1,\ldots,u_n)$ could be $n$ dimensional in which case $u_x$ denotes the collection of all element-wise first order derivatives $\frac{\partial u_1}{\partial x}, \ldots, \frac{\partial u_n}{\partial x}$. Similarly, $u_{xx}$ includes all element-wise second order derivatives $\frac{\partial^2 u_1}{\partial x^2}, \ldots, \frac{\partial^2 u_n}{\partial x^2}$.} Here, the subscripts denote partial differentiation in either time $t$ or space $x$.\footnote{ The space $x=(x_1,x_2,\ldots,x_m)$ could be a vector of dimension $m$. In this case, $u_x$ denotes the collection of all first order derivative $u_{x_1}, u_{x_2}, \ldots, u_{x_m}$ and $u_{xx}$ represents the set of all second order derivatives $u_{x_1 x_1}, u_{x_1 x_2},\ldots,u_{x_1 x_m}, \ldots, u_{x_m x_m}$.} Given a set of scattered and potentially noisy observations of the solution $u$, we are interested in learning the nonlinear function $\mathcal{N}$ and consequently discovering the hidden laws of physics that govern the evolution of the observed data.\\

For instance, let us assume that we would like to discover the Burger's equation \cite{basdevant1986spectral} in one space dimension $u_t = -u u_x + 0.1 u_{xx}$. Although not pursued in the current work, a viable approach \cite{Rudye1602614} to tackle this problem is to create a dictionary of possible terms and write the following expansion
\begin{eqnarray*}
\mathcal{N}(t,x,u,u_x,u_{xx},\ldots) &=& \alpha_{0,0} + \alpha_{1,0} u + \alpha_{2,0} u^2 + \alpha_{3,0} u^3 +\\
&~& \alpha_{0,1}u_{x} + \alpha_{1,1} u u_{x} + \alpha_{2,1} u^2 u_{x} + \alpha_{3,1} u^3 u_{x} + \\
&~& \alpha_{0,2}u_{xx} + \alpha_{1,2} u u_{xx} + \alpha_{2,2} u^2 u_{xx} + \alpha_{3,2} u^3 u_{xx} +\\
&~& \alpha_{0,3}u_{xxx} + \alpha_{1,3} u u_{xxx} + \alpha_{2,3} u^2 u_{xxx} + \alpha_{3,3} u^3 u_{xxx}.
\end{eqnarray*}
Given the aforementioned large collection of candidate terms for constructing the partial differential equation, one could then use sparse regression techniques \cite{Rudye1602614} to determine the coefficients $\alpha_{i,j}$ and consequently the right-hand-side terms that are contributing to the dynamics. A huge advantage of this approach is the interpretability of the learned equations. However, there are two major drawbacks with this method.\\

First, it relies on numerical differentiation to compute the derivatives involved in equation \eqref{eq:PDE}. Derivatives are taken either using finite differences for clean data or with polynomial interpolation in the presence of noise. Numerical approximations of derivatives are inherently ill-conditioned and unstable \cite{baydin2015automatic} even in the absence of noise in the data. This is due to the introduction of truncation and round-off errors inflicted by the limited precision of computations and the chosen value of the step size for finite differencing. Thus, this approach requires far more data points than library functions. This need for using a large number of points lies more in the numerical evaluation of derivatives than in supplying sufficient data for the regression.\\

Second, in applying the algorithm \cite{Rudye1602614} outlined above we assume that the chosen library is sufficiently rich to have a sparse representation of the time dynamics of the dataset. However, when applying this approach to a dataset where the dynamics are in fact unknown it is not unlikely that the basis chosen above is insufficient. Specially, in higher dimensions (i.e., for input $x$ or output $u$) the required number of terms to include in the library increases exponentially. Moreover, an additional issue with this approach is that it can only estimate parameters appearing as coefficients. For example, this method cannot estimate parameters of a partial differential equation (e.g., the sine-Gordon equation) involving a term like $\sin(\alpha u(x))$ with $\alpha$ being the unknown parameter, even if we include sines and cosines in the dictionary of possible terms.\\

One could avoid the first drawback concerning numerical differentiation by assigning prior distributions in the forms of Gaussian processes \cite{raissi2017machine,raissi2017hidden,raissi2017inferring,raissi2017numerical} or neural networks \cite{raissi2017physics_II,raissi2017physics_I} to the unknown solution $u$. Derivatives of the prior on $u$ can now be evaluated at machine precision using symbolic or automatic differentiation \cite{baydin2015automatic}. This removes the requirement for having or generating data on derivatives of the solution $u$. This is enabling as it allows us to work with noisy observations of the solution $u$, scattered in space and time. Moreover, this approach requires far fewer data points than the method proposed in \cite{Rudye1602614} simply because, as explained above, the need for using a large number of data points was due to the numerical evaluation of derivatives.\\

The second drawback can be addressed in a similar fashion by approximating the nonlinear function $\mathcal{N}$ (see equation \eqref{eq:PDE}) with a neural network. Representing the nonlinear function $\mathcal{N}$ by a deep neural network is the novelty of the current work. Deep neural networks are a richer family of function approximators and consequently we do not have to commit to a particular class of basis functions such as polynomials or sines and cosines.  This expressiveness comes at the cost of losing interpretability of the learned dynamics.  However, there is nothing hindering the use of a particular class of basis functions in order obtain more interpretable equations \cite{raissi2017physics_II}.

\section{Solution methodology}
We proceed by approximating both the solution $u$ and the nonlinear function $\mathcal{N}$ with two deep neural networks\footnote{ Representing the solution $u$ by a deep neural network is inspired by recent developments in \emph{physics-informed deep learning} \cite{raissi2017physics_I,raissi2017physics_II}, while approximating the nonlinear function $\mathcal{N}$ by another network is the novelty of this work.} and define a \emph{deep hidden physics model} $f$ to be given by
\begin{equation}
f := u_t - \mathcal{N}(t,x,u,u_x,u_{xx},\ldots).\label{eq:DeepHPM}
\end{equation}
We obtain the derivatives of the neural network $u$ with respect to time $t$ and space $x$ by applying the chain rule for differentiating compositions of functions using automatic differentiation \cite{baydin2015automatic}. It is worth emphasizing that automatic differentiation is different from, and in several aspects superior to, numerical or symbolic differentiation; two commonly encountered techniques of computing derivatives. In its most basic description \cite{baydin2015automatic}, automatic differentiation relies on the fact that all numerical computations are ultimately compositions of a finite set of elementary operations for which derivatives are known. Combining the derivatives of the constituent operations through the chain rule gives the derivative of the overall composition. This allows accurate evaluation of derivatives at machine precision with ideal asymptotic efficiency and only a small constant factor of overhead. In particular, to compute the derivatives involved in definition \eqref{eq:DeepHPM} we rely on Tensorflow \cite{abadi2016tensorflow} which is a popular and relatively well documented open source software library for automatic differentiation and deep learning computations.\\

Parameters of the neural networks $u$ and $\mathcal{N}$ can be learned by minimizing the sum of squared errors
\begin{equation}\label{eq:SSE}
\sum_{i=1}^{N} \left(|u(t^i,x^i) - u^i|^2 + |f(t^i,x^i)|^2\right),
\end{equation}
where $\{t^i, x^i, u^i\}_{i=1}^{N}$ denote the training data on $u$. The term $|u(t^i,x^i) - u^i|^2$ tries to fit the data by adjusting the parameters of the neural network $u$ while the term $|f(t^i,x^i)|^2$ learns the parameters of the network $\mathcal{N}$ by trying to satisfy the partial differential equation \eqref{eq:PDE} at the collocation points $(t^i,x^i)$. Training the parameters of the neural networks $u$ and $\mathcal{N}$ can be performed simultaneously by minimizing the sum of squared error \eqref{eq:SSE} or in a sequential fashion by training $u$ first and $\mathcal{N}$ second.\\

How can we make sure that the algorithm presented above results in an acceptable function $\mathcal{N}$? One answer would be to solve the learned equations and compare the resulting solution to the solution of the exact partial differential equation. However, it should be pointed out that the learned function $\mathcal{N}$ is a \emph{black-box} function; i.e., we do not know its functional form. Consequently, none of the classical partial differential equation solvers such as finite differences, finite elements or spectral methods are applicable here. Therefore, to solve the learned equations we have no other choice than to resort to modern black-box solvers such as \emph{physic informed neural networks} (PINNs) introduced in \cite{raissi2017physics_I}. The steps involved in PINNs as solvers\footnote{ PINNs have also been used in \cite{raissi2017physics_II} to address the problem of data-driven discovery of partial differential equations in cases where the nonlinear function $\mathcal{N}$ is known up to a set of parameters.} are similar to equations \eqref{eq:PDE}, \eqref{eq:DeepHPM}, and \eqref{eq:SSE} with the nonlinear function $\mathcal{N}$ being known and the data residing on the boundary of the domain.

\section{Results}
The proposed framework provides a universal treatment of nonlinear partial differential equations of fundamentally different nature. This generality will be demonstrated by applying the algorithm to a wide range of canonical problems spanning a number of scientific domains including the Burgers', Korteweg-de Vries (KdV), Kuramoto-Sivashinsky, nonlinear Schr\"{o}dinger, and Navier-Stokes equations. These examples are motivated by the pioneering work of \cite{Rudye1602614}. All data and codes used in this manuscript are publicly available on GitHub at \url{https://github.com/maziarraissi/DeepHPMs}.

\subsection{Burgers' equation}\label{sec:Burgers}
\begin{figure}[!t]
\includegraphics[width = 1.0\textwidth]{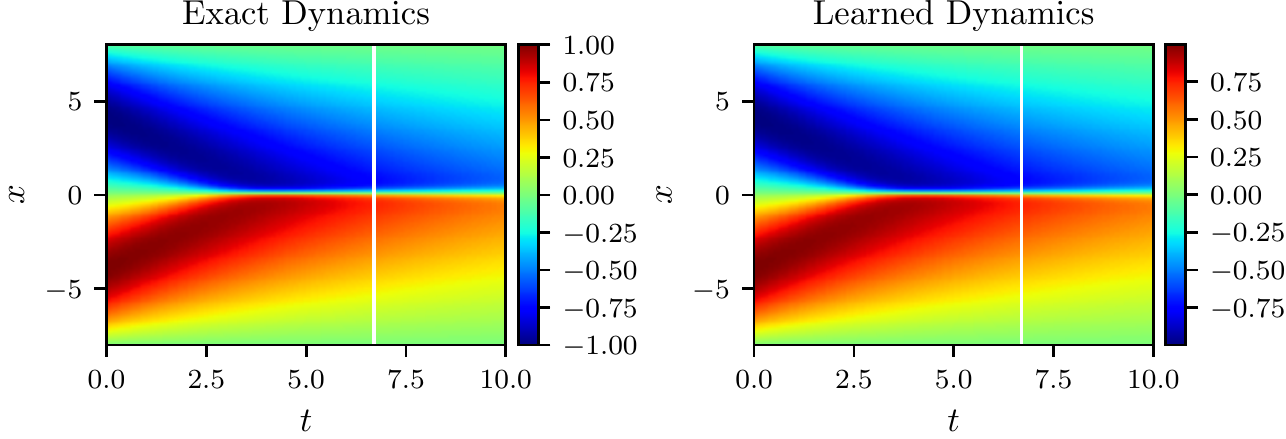}
\caption{{\em Burgers equation:} A solution to the Burger's equation (left panel) is compared to the corresponding solution of the learned partial differential equation (right panel). The identified system correctly captures the form of the dynamics and accurately reproduces the solution with a relative $L^2$-error of 4.78e-03. It should be emphasized that the training data are collected in roughly two-thirds of the domain between times $t = 0$ and $t=6.7$ represented by the white vertical lines. The algorithm is thus extrapolating from time $t=6.7$ onwards. The relative $L^2$-error on the training portion of the domain is 3.89e-03.}
\label{fig:Burgers}
\end{figure}

Let us start with the Burgers' equation arising in various areas of engineering and applied mathematics, including fluid mechanics, nonlinear acoustics, gas dynamics, and traffic flow \cite{basdevant1986spectral}. In one space dimension the Burgers' equation reads as
\begin{equation}\label{eq:Burgers}
u_t = - u u_x + 0.1 u_{xx}.
\end{equation}
To obtain a set of training and test data we simulate the Burger's equation \eqref{eq:Burgers} using conventional spectral methods. Specifically, starting from an initial condition $u(0,x) = -\sin(\pi x/8), ~x\in [-8,8]$ and assuming periodic boundary conditions, we integrate equation \eqref{eq:Burgers} up to the final time $t=10$. We use the Chebfun package \cite{driscoll2014chebfun} with a spectral Fourier discretization with 256 modes and a fourth-order explicit Runge-Kutta temporal integrator with time-step size $10^{-4}$. The solution is saved every $\Delta t = 0.05$ to give us a total of 201 snapshots. Out of this data-set, we generate a smaller training subset, scattered in space and time, by randomly sub-sampling $10000$ data points from time $t = 0$ to $t=6.7$. We call the portion of the domain from time $t = 0$ to $t = 6.7$ the training portion. The rest of the domain from time $t=6.7$ to the final time $t = 10$ will be referred to as the test portion. Using this terminology, we are in fact sub-sampling from the original dataset only in the training portion of the domain. Given the training data, we are interested in learning $\mathcal{N}$ as a function of the solution $u$ and its derivatives up to the 2nd order\footnote{ A detailed study of the choice of the order will be provided later in this section.}; i.e.,
\begin{equation}\label{eq:Burgers_learned}
u_t = \mathcal{N}(u,u_x,u_{xx}).
\end{equation}
We represent the solution $u$ by a 5-layer deep neural network with 50 neurons per hidden layer. Furthermore, we let $\mathcal{N}$ to be a neural network with 2 hidden layers and 100 neurons per hidden layer. These two networks are trained by minimizing the sum of squared errors loss of equation \eqref{eq:SSE}. To illustrate the effectiveness of our approach, we solve the learned partial differential equation \eqref{eq:Burgers_learned}, along with periodic boundary conditions and the same initial condition as the one used to generate the original dataset, using the PINNs algorithm \cite{raissi2017physics_I}. The original dataset alongside the resulting solution of the learned partial differential equation are depicted in figure \ref{fig:Burgers}. This figure indicates that our algorithm is able to accurately identify the underlying partial differential equation with a relative $L^2$-error of 4.78e-03. It should be highlighted that the training data are collected in roughly two-thirds of the domain between times $t = 0$ and $t=6.7$. The algorithm is thus extrapolating from time $t=6.7$ onwards. The relative $L^2$-error on the training portion of the domain is 3.89e-03.\\

Furthermore, we performed a systematic study of the reported results in figure \ref{fig:Burgers} with respect to noise levels in the data by keeping the total number of training observations as well as the neural network architectures fixed to the settings described above. In particular, we added white noise with magnitude equal to one, two, and five percent of the standard deviation of the solution function. The results of this study are summarized in table \ref{tab:Burgers}. The key observation here is that less noise in the data enhances the performance of the algorithm. Our experience so far indicates that the negative consequences of more noise in the data can be remedied to some extent by obtaining more data. Another fundamental point to make is that in general the choice of the neural network architectures, i.e., activation functions and number of layers/neurons, is crucial. However, in many cases of practical interest this choice still remains an art and systematic studies with respect to the neural network architectures often fail to reveal consistent patterns \cite{raissi2017physics_I,raissi2017physics_II,raissi2018multistep}. We usually observe some variability and non monotonic trends in systematic studies pertaining to the network architectures. In this regard, there exist a series of open questions mandating further investigations. For instance, how common techniques such as batch normalization, drop out, and $L^1$/$L^2$ regularization \cite{goodfellow2016deep} could enhance the robustness of the proposed algorithm with respect to the neural network architectures as well as noise in the data.\\

\begin{table}[!b]
\centering
\begin{tabular}{|c|c|c|c|c|}
\hline
                     & Clean data & 1\% noise & 2\% noise & 5\% noise \\
\hline
Relative $L^2$-error & 4.78e-03  & 2.64e-02  & 1.09e-01 & 4.56e-01\\
\hline
\end{tabular}
\caption{{\em Burgers' equation:} Relative $L^2$-error between solutions of the Burgers' equation and the learned partial differential equation as a function of noise corruption level in the data. Here, the total number of training data as well as the neural network architectures are kept fixed.} \label{tab:Burgers}
\end{table}

\begin{figure}[!t]
\includegraphics[width = 1.0\textwidth]{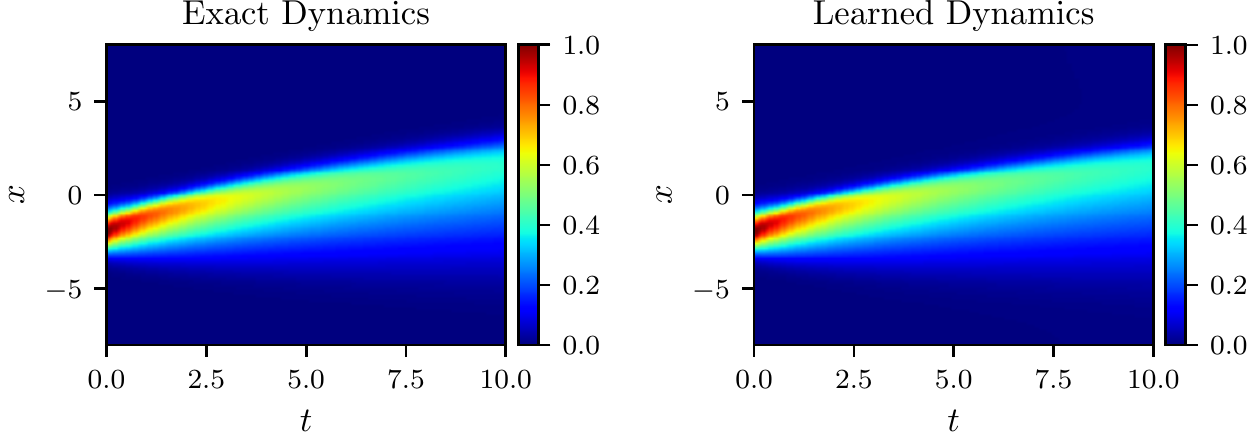}
\caption{{\em Burgers equation:} A solution to the Burger's equation (left panel) is compared to the corresponding solution of the learned partial differential equation (right panel). The identified system correctly captures the form of the dynamics and accurately reproduces the solution with a relative $L^2$-error of 7.33e-02. It should be highlighted that the algorithm is trained on a dataset totally different from the one used in this figure.}
\label{fig:Burgers_Extrapolate}
\end{figure}

To further scrutinize the performance of the algorithm, let us change the initial condition to $-\exp(-(x+2)^2)$ and solve the Burgers' equation \eqref{eq:Burgers} using a classical partial differential equation solver. In particular, the new data-set \cite{Rudye1602614} contains 101 time snapshots of a solution to the Burgers' equation \eqref{eq:Burgers} with a Gaussian initial condition, propagating into a traveling wave. The snapshots are $\Delta t = 0.1$ apart and stretch from time $t=0$ to $t=10$. The spatial discretization of each snapshot involves a uniform grid with 256 cells. We compare the resulting solution to the one obtained by solving the learned partial differential equation \eqref{eq:Burgers_learned} using the PINNs algorithm \cite{raissi2017physics_I}. It is worth emphasizing that the algorithm is trained on the dataset depicted in figure \ref{fig:Burgers} and is being tested on a totally different dataset as shown in figure \ref{fig:Burgers_Extrapolate}. The surprising result reported in figure \ref{fig:Burgers_Extrapolate} is a strong indication that the algorithm is capable of accurately identifying the underlying partial differential equation. The algorithm hasn't seen even a single observation of the dataset shown in figure \ref{fig:Burgers_Extrapolate} during model training and is yet capable of achieving a relatively accurate approximation of the true solution. The identified system reproduces the solution to the Burgers' equation with a relative $L^2$-error of 7.33e-02.\\

However, the aforementioned result seems to be sensitive to making the nonlinear function $\mathcal{N}$ of equation \eqref{eq:Burgers_learned} depend on either time $t$ or space $x$, or both. For instance, if we look for equations of the form $u_t = \mathcal{N}(x,u,u_x,u_{xx})$, the relative $L^2$-error between the exact and the learned solutions corresponding to figure \ref{fig:Burgers_Extrapolate} increases to 4.25e-01. Similarly, if we look for equations of the form $u_t = \mathcal{N}(t,u,u_x,u_{xx})$, the relative $L^2$-error increases to 2.58e-01. Moreover, if we make the nonlinear function $\mathcal{N}$ both time and space dependent the relative $L^2$-error increases even further to 1.46e+00. A possible explanation for this behavior could be that some of the dynamics in the training data is now being explained by time $t$ or space $x$. This makes the algorithm over-fit to the training data and consequently harder for it to generalize to datasets it has never seen before. Based on our experience more data seems to help resolve this issue. Another interesting observation is that if we train on the dataset presented in figure \ref{fig:Burgers_Extrapolate} and test on the one depicted in figure \ref{fig:Burgers}, i.e., swap the roles of these two datasets, the algorithm fails to generalize to the new test data. This observation suggests that the dynamics (see figure \ref{fig:Burgers}) generated by $-\sin(\pi x/8)$ as the initial condition is a reacher one compared to the dynamics (see figure \ref{fig:Burgers_Extrapolate}) generated by $\exp(-(x+2)^2)$. This is also evident from a visual comparison of the datasets given in figures \ref{fig:Burgers} and \ref{fig:Burgers_Extrapolate}.\\

Let us now take a closer look at equation \eqref{eq:Burgers_learned} and ask: what would happen if we included derivatives of higher order than two in our formulation? As will be demonstrated in the following, the algorithm proposed in the current work is capable of handling such cases, however, this is a fundamental question worthy of a moment of reflection. The choice of the order of the partial differential equation \eqref{eq:Burgers_learned} determines the form and the number of boundary conditions needed to end up with a well-posed problem. For instance, in the one dimensional setting of equation \eqref{eq:Burgers_learned} including a third order derivative requires three boundary conditions, namely $u(t,-8) = u(t,8)$, $u_x(t,-8) = u_x(t,8)$, and $u_{xx}(t,-8) = u_{xx}(t,8)$, assuming period boundary conditions. Consequently, in cases of practical interest, the available information on the boundary of the domain could help us determine the order of the partial differential equation we are try to identify. With this in mind, let us study the robustness of our framework with respect to the highest order of the derivatives included in equation \eqref{eq:Burgers_learned}. As for the boundary conditions, we use $u(t,-8) = u(t,8)$ and $u_x(t,-8) = u_x(t,8)$ when solving the identified partial differential equation regardless of the initial choice of its order. The results are summarized in table \ref{tab:Burgers_order}. The first column of table \ref{tab:Burgers_order} demonstrates that a single first order derivative is clearly not enough to capture the second order dynamics of the Burgers' equation. Moreover, the method seems to be generally robust with respect to the number and order of derivatives included in the nonlinear function $\mathcal{N}$. Therefore, in addition to any information residing on the domain boundary, studies such as table \ref{tab:Burgers_order}, albeit for training or validation datasets, could help us choose the best order for the underlying partial differential equation. In this case, table \ref{tab:Burgers_order} suggests the order of the equation to be two.\\

In addition, it must be stated that including higher order derivatives comes at the cost of reducing the speed of the algorithm due to the growth in the complexity of the resulting computational graph for the corresponding \emph{deep hidden physics model} (see equation \eqref{eq:DeepHPM}). Also, another drawback is that higher order derivatives are usually less accurate specially if one uses single precision floating-point system (float32) rather than double precision (float64). It is true that automatic differentiation enables us to evaluate derivatives at machine precision, however, for float32 the machine epsilon is approximately 1.19e-07. For improved performance in terms of speed of the algorithm and constrained by usual GPU (graphics processing unit) platforms we often end up using float32 which boils down to committing an error of approximately 1.19e-07 while computing the required derivatives. The aforementioned issues do not seem to be too serious since computer infrastructure (both hardware and software) for deep learning is constantly improving.

\begin{table}[!b]
\centering
\begin{tabular}{|c|c|c|c|c|}
\hline
                     & 1st order & 2nd order & 3rd order & 4th order \\
\hline
Relative $L^2$-error & 1.14e+00 & 1.29e-02  & 3.42e-02 & 5.54e-02 \\
\hline
\end{tabular}
\caption{{\em Burgers' equation:} Relative $L^2$-error between solutions of the Burgers' equation and the learned partial differential equation as a function of the highest order of spatial derivatives included in our formulation. For instance, the case corresponding to the 3rd order means that we are looking for a nonlinear function $\mathcal{N}$ such that $u_t = \mathcal{N}(u,u_x,u_{xx},u_{xxx})$. Here, the total number of training data as well as the neural network architectures are kept fixed and the data are assumed to be noiseless.}\label{tab:Burgers_order}
\end{table}

\subsection{The KdV equation}
\begin{figure}[!t]
\includegraphics[width = 1.0\textwidth]{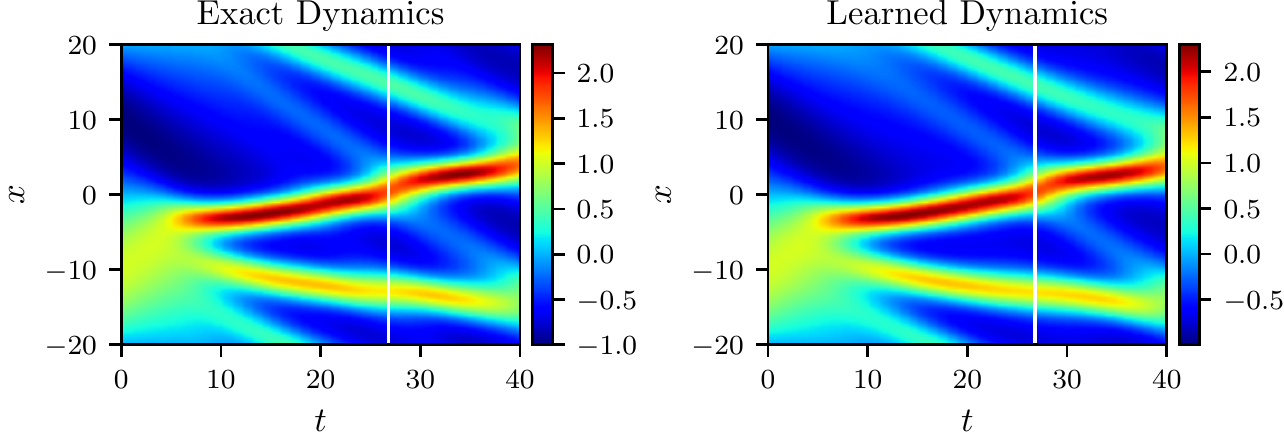}
\caption{{\em The KdV equation:} A solution to the KdV equation (left panel) is compared to the corresponding solution of the learned partial differential equation (right panel). The identified system correctly captures the form of the dynamics and accurately reproduces the solution with a relative $L^2$-error of 6.28e-02. It should be emphasized that the training data are collected in roughly two-thirds of the domain between times $t = 0$ and $t=26.8$ represented by the white vertical lines. The algorithm is thus extrapolating from time $t=26.8$ onwards. The relative $L^2$-error on the training portion of the domain is 3.78e-02.}
\label{fig:KdV}
\end{figure}

As a mathematical model of waves on shallow water surfaces one could consider the Korteweg-de Vries (KdV) equation. The KdV equation reads as
\begin{equation}\label{eq:KdV}
u_t = - u u_x - u_{xxx}.
\end{equation}
To obtain a set of training data we simulate the KdV equation \eqref{eq:KdV} using conventional spectral methods. In particular, we start from an initial condition $u(0,x) = -\sin(\pi x/20), ~x\in [-20,20]$ and assume periodic boundary conditions. We integrate equation \eqref{eq:KdV} up to the final time $t = 40$. We use the Chebfun package \cite{driscoll2014chebfun} with a spectral Fourier discretization with 512 modes and a fourth-order explicit Runge-Kutta temporal integrator with time-step size $10^{-4}$. The solution is saved every $\Delta t = 0.2$ to give us a total of 201 snapshots. Out of this data-set, we generate a smaller training subset, scattered in space and time, by randomly sub-sampling $10000$ data points from time $t = 0$ to $t=26.8$. In other words, we are sub-sampling from the original dataset only in the training portion of the domain from time $t = 0$ to $t=26.8$. Given the training data, we are interested in learning $\mathcal{N}$ as a function of the solution $u$ and its derivatives up to the 3rd order\footnote{ A detailed study of the choice of the order is provided in section \ref{sec:Burgers} for the Burgers' equation.}; i.e.,
\begin{equation}\label{eq:KdV_learned}
u_t = \mathcal{N}(u,u_x,u_{xx},u_{xxx}).
\end{equation}
We represent the solution $u$ by a 5-layer deep neural network with 50 neurons per hidden layer. Furthermore, we let $\mathcal{N}$ to be a neural network with 2 hidden layers and 100 neurons per hidden layer. These two networks are trained by minimizing the sum of squared errors loss of equation \eqref{eq:SSE}. To illustrate the effectiveness of our approach, we solve the learned partial differential equation \eqref{eq:KdV_learned} using the PINNs algorithm \cite{raissi2017physics_I}. We assume periodic boundary conditions and the same initial condition as the one used to generate the original dataset. The resulting solution of the learned partial differential equation as well as the exact solution of the KdV equation are depicted in figure \ref{fig:KdV}. This figure indicates that our algorithm is capable of accurately identifying the underlying partial differential equation with a relative $L^2$-error of 6.28e-02. It should be highlighted that the training data are collected in roughly two-thirds of the domain between times $t = 0$ and $t=26.8$. The algorithm is thus extrapolating from time $t=26.8$ onwards. The corresponding relative $L^2$-error on the training portion of the domain is 3.78e-02.\\

\begin{figure}[!t]
\includegraphics[width = 1.0\textwidth]{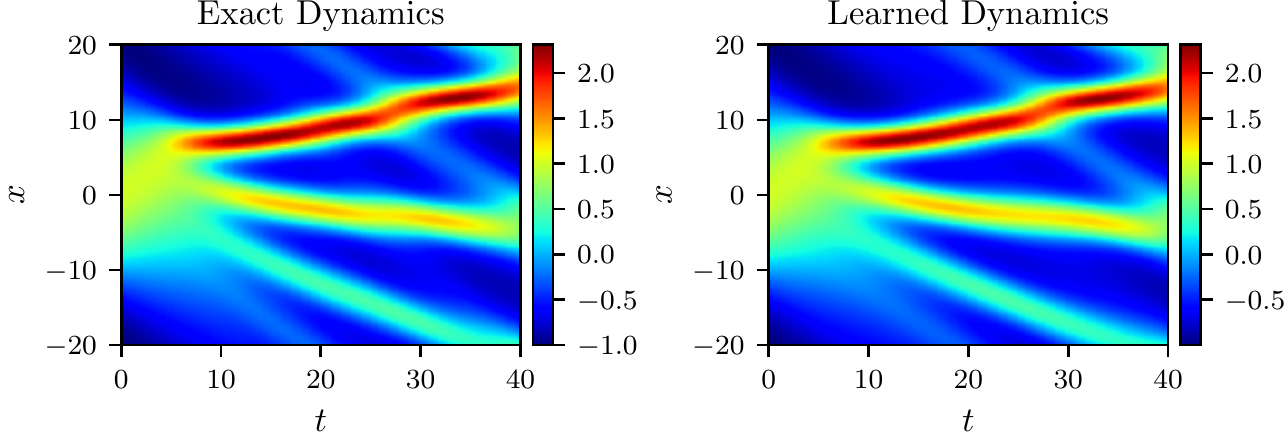}
\caption{{\em The KdV equation:} A solution to the KdV equation (left panel) is compared to the corresponding solution of the learned partial differential equation (right panel). The identified system correctly captures the form of the dynamics and accurately reproduces the solution with a relative $L^2$-error of 3.44e-02. It should be highlighted that the algorithm is trained on a dataset totally different from the one shown in this figure.}
\label{fig:KdV_Extrapolate}
\end{figure}

To test the algorithm even further, let us change the initial condition to $\cos(-\pi x/20)$ and solve the KdV \eqref{eq:KdV} using the conventional spectral method outlined above. We compare the resulting solution to the one obtained by solving the learned partial differential equation \eqref{eq:Burgers_learned} using the PINNs algorithm \cite{raissi2017physics_I}. It is worth emphasizing that the algorithm is trained on the dataset depicted in figure \ref{fig:KdV} and is being tested on a different dataset as shown in figure \ref{fig:KdV_Extrapolate}. The surprising result reported in figure \ref{fig:KdV_Extrapolate} strongly indicates that the algorithm is accurately learning the underlying partial differential equation; i.e., the nonlinear function $\mathcal{N}$. The algorithm hasn't seen the dataset shown in figure \ref{fig:KdV_Extrapolate} during model training and is yet capable of achieving a relatively accurate approximation of the true solution. To be precise, the identified system reproduces the solution to the KdV equation with a relative $L^2$-error of 3.44e-02.

\subsection{Kuramoto-Sivashinsky equation}
\begin{figure}[!t]
\includegraphics[width = 1.0\textwidth]{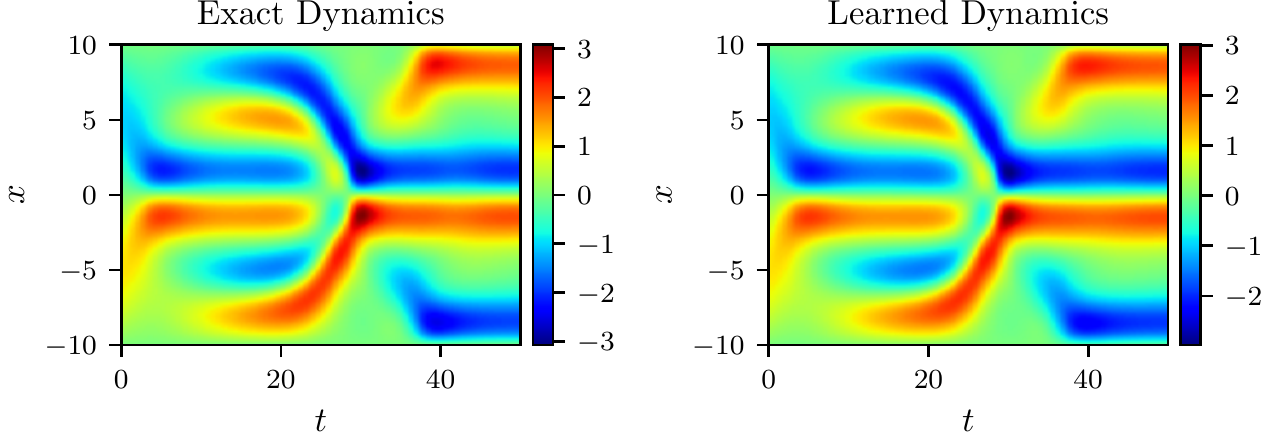}
\caption{{\em Kuramoto-Sivashinsky equation:} A solution to the Kuramoto-Sivashinsky equation (left panel) is compared to the corresponding solution of the learned partial differential equation (right panel). The identified system correctly captures the form of the dynamics and reproduces the solution with a relative $L^2$-error of 7.63e-02.}
\label{fig:KS}
\end{figure}

As a canonical model of a pattern forming system with spatio-temporal chaotic behavior we consider the Kuramoto-Sivashinsky equation. In one space dimension the Kuramoto-Sivashinsky equation reads as
\begin{equation}\label{eq:KS}
u_t = - u u_x - u_{xx} - u_{xxxx}.
\end{equation}
We generate a dataset containing a direct numerical solution of the Kuramoto-Sivashinsky \eqref{eq:KS} equation with 512 spatial points and 251 snapshots. To be precise, assuming periodic boundary conditions, we start from the initial condition $u(0,x) = -\sin(\pi x/10), ~x\in [-10,10]$ and integrate equation \eqref{eq:KS} up to the final time $t = 50$. We employ the Chebfun package \cite{driscoll2014chebfun} with a spectral Fourier discretization with 512 modes and a fourth-order explicit Runge-Kutta temporal integrator with time-step size $10^{-5}$. The snapshots are saved every $\Delta t = 0.2$. From this dataset, we create a smaller training subset, scattered in space and time, by randomly sub-sampling $10000$ data points from time $t = 0$ to the final time $t=50.0$. Given the resulting training data, we are interested in learning $\mathcal{N}$ as a function of the solution $u$ and its derivatives up to the 4rd order\footnote{ A detailed study of the choice of the order is provided in section \ref{sec:Burgers} for the Burgers' equation.}; i.e.,
\begin{equation}\label{eq:KS_learned}
u_t = \mathcal{N}(u,u_x,u_{xx},u_{xxx},u_{xxxx}).
\end{equation}
We let the solution $u$ to be represented by a 5-layer deep neural network with 50 neurons per hidden layer. Furthermore, we approximate the nonlinear function $\mathcal{N}$ by a neural network with 2 hidden layers and 100 neurons per hidden layer. These two networks are trained by minimizing the sum of squared errors loss of equation \eqref{eq:SSE}. To demonstrate the effectiveness of our approach, we solve the learned partial differential equation \eqref{eq:KS_learned} using the PINNs algorithm \cite{raissi2017physics_I}. We assume the same initial and boundary conditions as the ones used to generate the original dataset. The resulting solution of the learned partial differential equation alongside the exact solution of the Kuramoto-Sivashinsky equation are depicted in figure \ref{fig:KS}. This figure indicates that our algorithm is capable of identifying the underlying partial differential equation with a relative $L^2$-error of 7.63e-02.\\

\begin{figure}[!t]
\includegraphics[width = 1.0\textwidth]{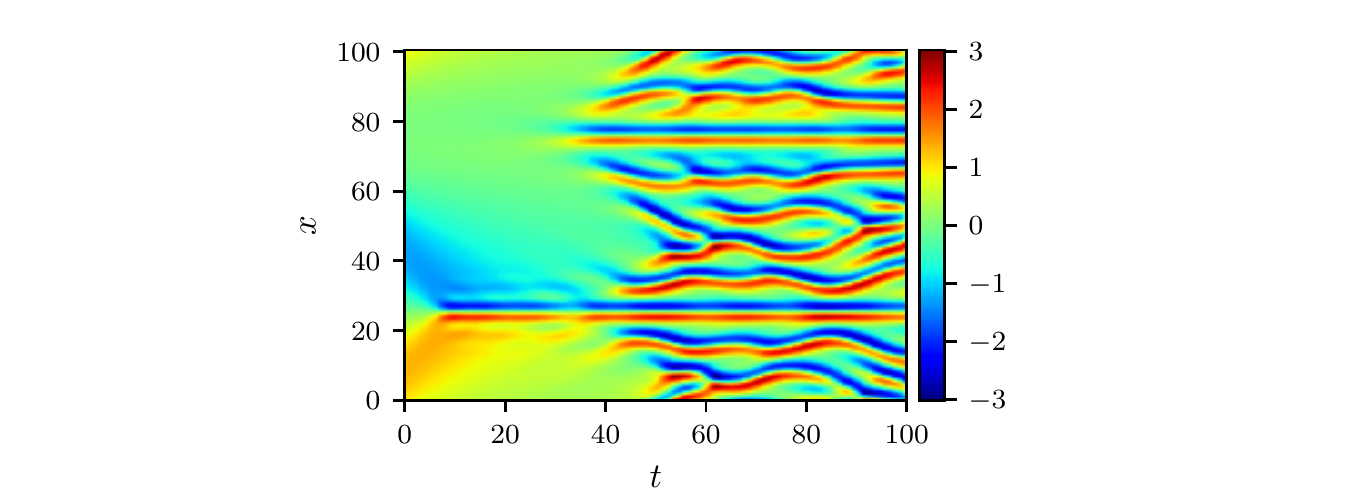}
\caption{{\em Kuramoto-Sivashinsky equation:} A solution to the Kuramoto-Sivashinsky equation. This problem remains stubbornly unsolved in the face of the algorithm proposed in the current work.}
\label{fig:KS_nasty}
\end{figure}

However, it must be mentioned that we are avoiding the regimes where the Kuramoto-Sivashinsky equation becomes chaotic. For instance, by changing the domain to $[0,32\pi]$ and the initial condition to $\cos(x/16)(1+\sin(x/16))$ and by integrating the Kuramoto-Sivashinsky equation \eqref{eq:KS} up to the final time $t = 100$, one could end up with a relatively complicated solution as depicted in figure \ref{fig:KS_nasty}. This problem remains stubbornly unsolved in the face of the algorithm proposed in the current work as well as the PINNs framework introduced in \cite{raissi2017physics_I,raissi2017physics_II}. It is not difficult for a plain vanilla neural network to approximate the function depicted in figure \ref{fig:KS_nasty}. However, as we compute the derivatives required in equation \eqref{eq:DeepHPM}, minimizing the loss function \eqref{eq:SSE} becomes a challenge. Understanding what makes this problem hard to solve should be at the core of future extensions of this line of research.

\subsection{Nonlinear Schr\"{o}dinger equation}
\begin{figure}[!t]
\includegraphics[width = 1.0\textwidth]{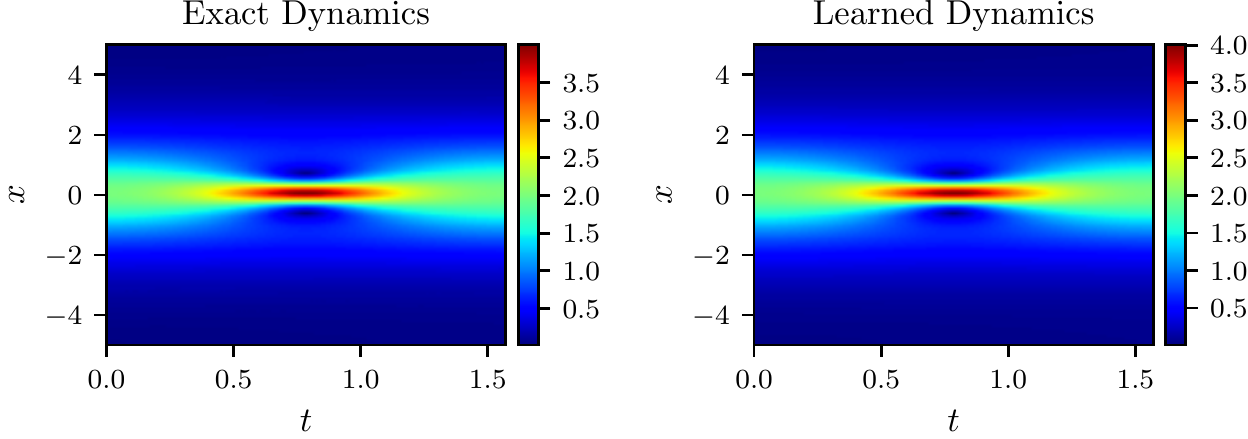}
\caption{{\em Nonlinear Schr\"{o}dinger equation:} Absolute value of a solution to the nonlinear Schr\"{o}dinger equation (left panel) is compared to the absolute value of the corresponding solution of the learned partial differential equation (right panel). The identified system correctly captures the form of the dynamics and accurately reproduces the absolute value of the solution with a relative $L^2$-error of 6.28e-03.}
\label{fig:NLS}
\end{figure}

The one-dimensional nonlinear Schr\"{o}dinger equation is a classical field equation that is used to study nonlinear wave propagation in optical fibers and/or waveguides, Bose-Einstein condensates, and plasma waves. This example aims to highlight the ability of our framework to handle complex-valued solutions as well as different types of nonlinearities in the governing partial differential equations. The nonlinear Schr\"{o}dinger equation is given by
\begin{equation}\label{eq:Schrodinger}
\psi_t  = 0.5 i \psi_{xx} + i |\psi|^2 \psi.
\end{equation}
Let $u$ denote the real part of $\psi$ and $v$ the imaginary part. Then, the nonlinear Schr\"{o}dinger equation can be equivalently written as a system of partial differential equations
\begin{equation}\label{eq:SchrodingerSystem}
\begin{array}{l}
u_t = - 0.5 v_{xx} - (u^2 + v^2) v,\\
v_t = 0.5 u_{xx} + (u^2 + v^2) u.
\end{array}
\end{equation}
In order to assess the performance of our method, we simulate equation \eqref{eq:Schrodinger} using conventional spectral methods to create a high-resolution data set. Specifically, starting from an initial state $\psi(0,x) = 2\ \text{sech}(x)$ and assuming periodic boundary conditions $\psi(t,-5) = \psi(t,5)$ and $\psi_x(t,-5) = \psi_x(t,5)$, we integrate equation \eqref{eq:Schrodinger} up to a final time $t=\pi/2$ using the Chebfun package \cite{driscoll2014chebfun}. We are in fact using a spectral Fourier discretization with 256 modes and a fourth-order explicit Runge-Kutta temporal integrator with time-step $\Delta{t} = \pi/2 \cdot 10^{-6}$. The solution is saved approximately every $\Delta t = 0.0031$ to give us a total of 501 snapshots. Out of this data-set, we generate a smaller training subset, scattered in space and time, by randomly sub-sampling $10000$ data points from time $t = 0$ up to the final time $t=\pi/2$. Given the resulting training data, we are interested in learning two nonlinear functions $\mathcal{N}_1$ and $\mathcal{N}_2$ as functions of the solutions $u, v$ and their derivatives up to the 2nd order\footnote{ A detailed study of the choice of the order is provided in section \ref{sec:Burgers} for the Burgers' equation.}; i.e.,
\begin{equation}\label{eq:Schrodinger_learned}
\begin{array}{l}
u_t = \mathcal{N}_1(u,v,u_x,v_x,u_{xx},v_{xx}),\\
v_t = \mathcal{N}_2(u,v,u_x,v_x,u_{xx},v_{xx}).
\end{array}
\end{equation}
We represent the solutions $u$ and $v$ by two 5-layer deep neural networks with 50 neurons per hidden layer. Furthermore, we let $\mathcal{N}_1$ and $\mathcal{N}_2$ to be two neural networks with 2 hidden layers and 100 neurons per hidden layer. These four networks are trained by minimizing the sum of squared errors loss of equation \eqref{eq:SSE}. To illustrate the effectiveness of our approach, we solve the learned partial differential equation \eqref{eq:Schrodinger_learned}, along with periodic boundary conditions and the same initial condition as the one used to generate the original dataset, using the PINNs algorithm \cite{raissi2017physics_I}. The original dataset (in absolute values, i.e., $|\psi| = \sqrt{u^2+v^2}$) alongside the resulting solution (also in absolute values) of the learned partial differential equation are depicted in figure \ref{fig:NLS}. This figure indicates that our algorithm is able to accurately identify the underlying partial differential equation with a relative $L^2$-error of 6.28e-03.\\

\subsection{Navier-Stokes equation}
\begin{figure}[!t]
\includegraphics[width = 1.0\textwidth]{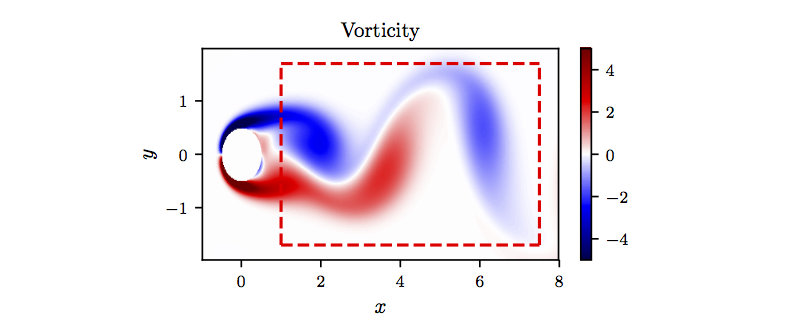}
\caption{{\em Navier-Stokes equation:} A snapshot of the vorticity field of a solution to the Navier-Stokes equations for the fluid flow past a cylinder. The dashed red box in this panel specifies the sampling region.}
\label{fig:Cylinder_vorticity}
\end{figure}

Let us consider the Navier-Stokes equation in two dimensions\footnote{ It is straightforward to generalize the proposed framework to the Navier-Stokes equation in three dimensions (3D).} (2D) given explicitly by
\begin{equation}\label{eq:NavierStokes}
\begin{array}{c}
w_t  = - u w_x - v w_y + 0.01(w_{xx} + w_{yy}),
\end{array}
\end{equation}
where $w$ denotes the vorticity, $u$ the $x$-component of the velocity field, and $v$ the $y$-component. To generate a training dataset for this problem we follow the exact same instructions as the ones provided in \cite{kutz2016dynamic, Rudye1602614}. Specifically, we simulate the Navier-Stokes equations describing the two-dimensional fluid flow past a circular cylinder at Reynolds number 100 using the Immersed Boundary Projection Method \cite{taira2007immersed, colonius2008fast}. This approach utilizes a multi-domain scheme with four nested domains, each successive grid being twice as large as the previous one. Length and time are nondimensionalized so that the cylinder has unit diameter and the flow has unit velocity. Data is collected on the finest domain with dimensions $9 \times 4$ at a grid resolution of $449 \times 199$. The flow solver uses a 3rd-order Runge Kutta integration scheme with a time step of t = 0.02, which has been verified to yield well-resolved and converged flow fields. After simulation converges to steady periodic vortex shedding, $151$ flow snapshots are saved every $\Delta t = 0.02$. We use a small portion of the resulting data-set for model training. In particular, we subsample $50000$ data points, scattered in space and time, in the rectangular region (dashed red box) downstream of the cylinder as shown in figure \ref{fig:Cylinder_vorticity}.\\

\begin{figure}[!t]
\includegraphics[width = 1.0\textwidth]{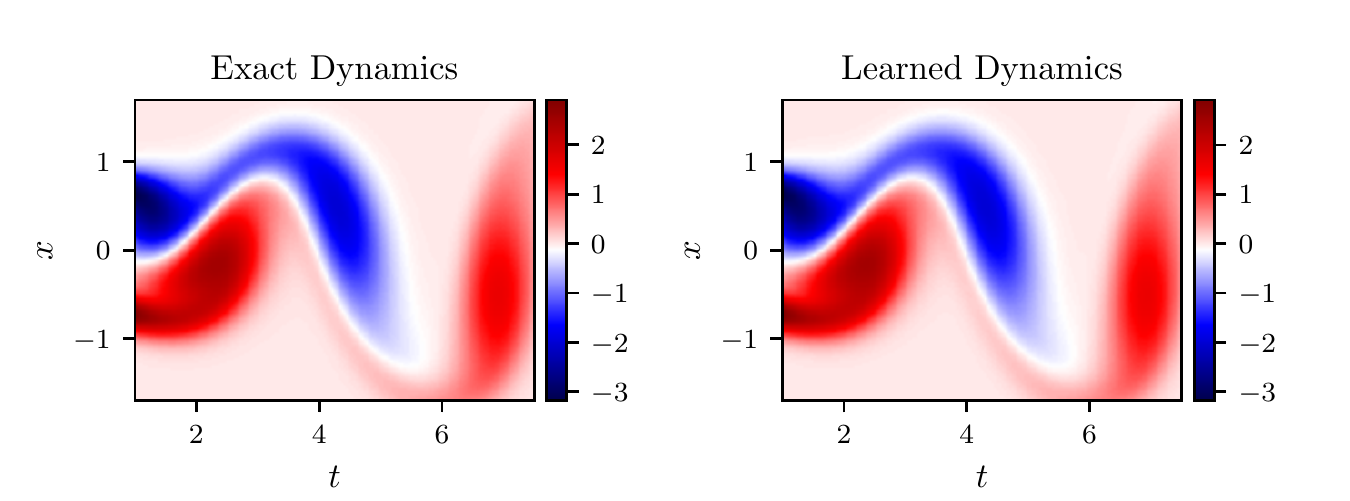}
\caption{{\em Navier-Stokes equation:} A randomly picked snapshot of a solution to the Navier-Stokes equation (left panel) is compared to the corresponding snapshot of the solution of the learned partial differential equation (right panel). The identified system correctly captures the form of the dynamics and accurately reproduces the solution with a relative $L^2$-error of 5.79e-03 in space and time.}
\label{fig:NavierStokes}
\end{figure}

Given the training data, we are interested in learning $\mathcal{N}$ as a function of the stream-wise $u$ and transverse $v$ velocity components in addition to the vorticity $w$ and its derivatives up to the 2nd order\footnote{ A detailed study of the choice of the order is provided in this section \ref{eq:Burgers} for the Burgers' equation.}; i.e.,
\begin{equation}\label{eq:NavierStokes_learned}
w_t = \mathcal{N}(u,v,w,w_x,w_{xx}).
\end{equation}
We represent the solution $w$ by a 5-layer deep neural network with 200 neurons per hidden layer. Furthermore, we let $\mathcal{N}$ to be a neural network with 2 hidden layers and 100 neurons per hidden layer. These two networks are trained by minimizing the sum of squared errors loss
\begin{equation*}
\sum_{i=1}^{N} \left(|w(t^i,x^i,y^i) - w^i|^2 + |f(t^i,x^i,y^i)|^2\right),
\end{equation*}
where $\{t^i, x^i, u^i, v^i, w^i\}_{i=1}^{N}$ denote the training data on $u$ and $f(t^i,x^i,y^i)$ is given by
\[
w_t(t^i,x^i,y^i) - \mathcal{N}(u^i,v^i,w(t^i,x^i,y^i),w_x(t^i,x^i,y^i),w_{xx}(t^i,x^i,y^i)).
\]
To illustrate the effectiveness of our approach, we solve the learned partial differential equation \eqref{eq:NavierStokes_learned}, in the region specified in figure \ref{fig:Cylinder_vorticity} by the dashed red box, using the PINNs algorithm \cite{raissi2017physics_I}. We use the exact solution to provide us with the required Dirichlet boundary conditions as well as the initial condition needed to solve the leaned partial differential equation \eqref{eq:NavierStokes_learned}. A randomly picked snapshot of the vorticity field in the original dataset alongside the corresponding snapshot of the solution of the learned partial differential equation are depicted in figure \ref{fig:NavierStokes}. This figure indicates that our algorithm is able to accurately identify the underlying partial differential equation with a relative $L^2$-error of 5.79e-03 in space and time.

\section{Summary and Discussion} 
We have presented a deep learning approach for extracting nonlinear partial differential equations from spatio-temporal datasets. The proposed algorithm leverages recent developments in automatic differentiation to construct efficient algorithms for learning infinite dimensional dynamical systems using deep neural networks. In order to validate the performance of our approach we had no other choice than to rely on black-box solvers (see e.g., \cite{raissi2017physics_II}). This signifies the importance of developing general purpose partial differential equation solvers. Developing these types of solvers is still in its infancy and more collaborative work is needed to bring them to the maturity level of conventional methods such as finite elements, finite differences, and spectral methods which have been around for more than half a century or so.\\

There exist a series of open questions mandating further investigations. For instance, many real-world partial differential equations depend on parameters and, when the parameters are varied, they may go through bifurcations (e.g., the Reynold number for the Navier-Stokes equations). Here, the goal would be to collect data from the underlying dynamics corresponding to various parameter values, and infer the parameterized partial differential equation. Another exciting avenue of future research would be to apply convolutional architectures \cite{goodfellow2016deep} for mitigating the complexity associated with partial differential equations with very high-dimensional inputs. These types of equations appear routinely in dynamic programming, optimal control, or reinforcement learning. Moreover, a quick look at the list of nonlinear partial differential equations on Wikipedia reveals that many of these equations take the form specified in equation \eqref{eq:PDE}. However, a handful of them do not take this form, including the Boussinesq type equation $u_{tt} - u_{xx} - 2 \alpha (u u_x)_x - \beta u_{xxtt} = 0$. It would be interesting to extend the framework outlined in the current work to incorporate all such cases. In the end, it is not always clear what measurements of a dynamical system to take. Even if we did know, collecting these measurements might be prohibitively expensive. It is well-known that time-delay coordinates of a single variable can act as additional variables. It might be interesting to investigating this idea for the infinite dimensional setting of partial differential equations.

\section*{Acknowledgements}
This work received support by the DARPA EQUiPS grant N66001-15-2-4055  and the AFOSR grant FA9550-17-1-0013. All data and codes used in this manuscript are publicly available on GitHub at \url{https://github.com/maziarraissi/DeepHPMs}.





\bibliographystyle{model1-num-names}
\bibliography{sample.bib}

\begin{thebibliography}{47}
\expandafter\ifx\csname natexlab\endcsname\relax\def\natexlab#1{#1}\fi
\providecommand{\bibinfo}[2]{#2}
\ifx\xfnm\relax \def\xfnm[#1]{\unskip,\space#1}\fi
\bibitem[{Rudy et~al.(2017)Rudy, Brunton, Proctor, and Kutz}]{Rudye1602614}
\bibinfo{author}{S.~H. Rudy}, \bibinfo{author}{S.~L. Brunton},
  \bibinfo{author}{J.~L. Proctor}, \bibinfo{author}{J.~N. Kutz},
\newblock \bibinfo{title}{Data-driven discovery of partial differential
  equations},
\newblock \bibinfo{journal}{Science Advances} \bibinfo{volume}{3}
  (\bibinfo{year}{2017}).
\bibitem[{Crutchfield and McNamara(1987)}]{crutchfield1987equations}
\bibinfo{author}{J.~P. Crutchfield}, \bibinfo{author}{B.~S. McNamara},
\newblock \bibinfo{title}{Equations of motion from a data series},
\newblock \bibinfo{journal}{Complex systems} \bibinfo{volume}{1}
  (\bibinfo{year}{1987}) \bibinfo{pages}{121}.
\bibitem[{Kevrekidis et~al.(2003)Kevrekidis, Gear, Hyman, Kevrekidid, Runborg,
  Theodoropoulos et~al.}]{kevrekidis2003equation}
\bibinfo{author}{I.~G. Kevrekidis}, \bibinfo{author}{C.~W. Gear},
  \bibinfo{author}{J.~M. Hyman}, \bibinfo{author}{P.~G. Kevrekidid},
  \bibinfo{author}{O.~Runborg}, \bibinfo{author}{C.~Theodoropoulos}, et~al.,
\newblock \bibinfo{title}{Equation-free, coarse-grained multiscale computation:
  Enabling mocroscopic simulators to perform system-level analysis},
\newblock \bibinfo{journal}{Communications in Mathematical Sciences}
  \bibinfo{volume}{1} (\bibinfo{year}{2003}) \bibinfo{pages}{715--762}.
\bibitem[{Raissi et~al.(2018)Raissi, Perdikaris, and
  Karniadakis}]{raissi2018multistep}
\bibinfo{author}{M.~Raissi}, \bibinfo{author}{P.~Perdikaris},
  \bibinfo{author}{G.~E. Karniadakis},
\newblock \bibinfo{title}{Multistep neural networks for data-driven discovery
  of nonlinear dynamical systems},
\newblock \bibinfo{journal}{arXiv preprint arXiv:1801.01236}
  (\bibinfo{year}{2018}).
\bibitem[{Gonzalez-Garcia et~al.(1998)Gonzalez-Garcia, Rico-Martinez, and
  Kevrekidis}]{gonzalez1998identification}
\bibinfo{author}{R.~Gonzalez-Garcia}, \bibinfo{author}{R.~Rico-Martinez},
  \bibinfo{author}{I.~Kevrekidis},
\newblock \bibinfo{title}{Identification of distributed parameter systems: A
  neural net based approach},
\newblock \bibinfo{journal}{Computers \& chemical engineering}
  \bibinfo{volume}{22} (\bibinfo{year}{1998}) \bibinfo{pages}{S965--S968}.
\bibitem[{Anderson et~al.(1996)Anderson, Kevrekidis, and
  Rico-Martinez}]{anderson1996comparison}
\bibinfo{author}{J.~Anderson}, \bibinfo{author}{I.~Kevrekidis},
  \bibinfo{author}{R.~Rico-Martinez},
\newblock \bibinfo{title}{A comparison of recurrent training algorithms for
  time series analysis and system identification},
\newblock \bibinfo{journal}{Computers \& chemical engineering}
  \bibinfo{volume}{20} (\bibinfo{year}{1996}) \bibinfo{pages}{S751--S756}.
\bibitem[{Rico-Martinez et~al.(1992)Rico-Martinez, Krischer, Kevrekidis, Kube,
  and Hudson}]{rico1992discrete}
\bibinfo{author}{R.~Rico-Martinez}, \bibinfo{author}{K.~Krischer},
  \bibinfo{author}{I.~Kevrekidis}, \bibinfo{author}{M.~Kube},
  \bibinfo{author}{J.~Hudson},
\newblock \bibinfo{title}{Discrete-vs. continuous-time nonlinear signal
  processing of cu electrodissolution data},
\newblock \bibinfo{journal}{Chemical Engineering Communications}
  \bibinfo{volume}{118} (\bibinfo{year}{1992}) \bibinfo{pages}{25--48}.
\bibitem[{Voss et~al.(1999)Voss, Kolodner, Abel, and
  Kurths}]{voss1999amplitude}
\bibinfo{author}{H.~U. Voss}, \bibinfo{author}{P.~Kolodner},
  \bibinfo{author}{M.~Abel}, \bibinfo{author}{J.~Kurths},
\newblock \bibinfo{title}{Amplitude equations from spatiotemporal binary-fluid
  convection data},
\newblock \bibinfo{journal}{Physical review letters} \bibinfo{volume}{83}
  (\bibinfo{year}{1999}) \bibinfo{pages}{3422}.
\bibitem[{Sugihara et~al.(2012)Sugihara, May, Ye, Hsieh, Deyle, Fogarty, and
  Munch}]{sugihara2012detecting}
\bibinfo{author}{G.~Sugihara}, \bibinfo{author}{R.~May},
  \bibinfo{author}{H.~Ye}, \bibinfo{author}{C.-h. Hsieh},
  \bibinfo{author}{E.~Deyle}, \bibinfo{author}{M.~Fogarty},
  \bibinfo{author}{S.~Munch},
\newblock \bibinfo{title}{Detecting causality in complex ecosystems},
\newblock \bibinfo{journal}{science} \bibinfo{volume}{338}
  (\bibinfo{year}{2012}) \bibinfo{pages}{496--500}.
\bibitem[{Ye et~al.(2015)Ye, Beamish, Glaser, Grant, Hsieh, Richards, Schnute,
  and Sugihara}]{ye2015equation}
\bibinfo{author}{H.~Ye}, \bibinfo{author}{R.~J. Beamish},
  \bibinfo{author}{S.~M. Glaser}, \bibinfo{author}{S.~C. Grant},
  \bibinfo{author}{C.-h. Hsieh}, \bibinfo{author}{L.~J. Richards},
  \bibinfo{author}{J.~T. Schnute}, \bibinfo{author}{G.~Sugihara},
\newblock \bibinfo{title}{Equation-free mechanistic ecosystem forecasting using
  empirical dynamic modeling},
\newblock \bibinfo{journal}{Proceedings of the National Academy of Sciences}
  \bibinfo{volume}{112} (\bibinfo{year}{2015}) \bibinfo{pages}{E1569--E1576}.
\bibitem[{Roberts(2014)}]{roberts2014model}
\bibinfo{author}{A.~J. Roberts}, \bibinfo{title}{Model emergent dynamics in
  complex systems}, \bibinfo{publisher}{SIAM}, \bibinfo{year}{2014}.
\bibitem[{Schmidt et~al.(2011)Schmidt, Vallabhajosyula, Jenkins, Hood, Soni,
  Wikswo, and Lipson}]{schmidt2011automated}
\bibinfo{author}{M.~D. Schmidt}, \bibinfo{author}{R.~R. Vallabhajosyula},
  \bibinfo{author}{J.~W. Jenkins}, \bibinfo{author}{J.~E. Hood},
  \bibinfo{author}{A.~S. Soni}, \bibinfo{author}{J.~P. Wikswo},
  \bibinfo{author}{H.~Lipson},
\newblock \bibinfo{title}{Automated refinement and inference of analytical
  models for metabolic networks},
\newblock \bibinfo{journal}{Physical biology} \bibinfo{volume}{8}
  (\bibinfo{year}{2011}) \bibinfo{pages}{055011}.
\bibitem[{Daniels and Nemenman(2015{\natexlab{a}})}]{daniels2015automated}
\bibinfo{author}{B.~C. Daniels}, \bibinfo{author}{I.~Nemenman},
\newblock \bibinfo{title}{Automated adaptive inference of phenomenological
  dynamical models},
\newblock \bibinfo{journal}{Nature communications} \bibinfo{volume}{6}
  (\bibinfo{year}{2015}{\natexlab{a}}).
\bibitem[{Daniels and Nemenman(2015{\natexlab{b}})}]{daniels2015efficient}
\bibinfo{author}{B.~C. Daniels}, \bibinfo{author}{I.~Nemenman},
\newblock \bibinfo{title}{Efficient inference of parsimonious phenomenological
  models of cellular dynamics using s-systems and alternating regression},
\newblock \bibinfo{journal}{PloS one} \bibinfo{volume}{10}
  (\bibinfo{year}{2015}{\natexlab{b}}) \bibinfo{pages}{e0119821}.
\bibitem[{Majda et~al.(2009)Majda, Franzke, and Crommelin}]{majda2009normal}
\bibinfo{author}{A.~J. Majda}, \bibinfo{author}{C.~Franzke},
  \bibinfo{author}{D.~Crommelin},
\newblock \bibinfo{title}{Normal forms for reduced stochastic climate models},
\newblock \bibinfo{journal}{Proceedings of the National Academy of Sciences}
  \bibinfo{volume}{106} (\bibinfo{year}{2009}) \bibinfo{pages}{3649--3653}.
\bibitem[{Giannakis and Majda(2012)}]{giannakis2012nonlinear}
\bibinfo{author}{D.~Giannakis}, \bibinfo{author}{A.~J. Majda},
\newblock \bibinfo{title}{Nonlinear laplacian spectral analysis for time series
  with intermittency and low-frequency variability},
\newblock \bibinfo{journal}{Proceedings of the National Academy of Sciences}
  \bibinfo{volume}{109} (\bibinfo{year}{2012}) \bibinfo{pages}{2222--2227}.
\bibitem[{Mezi{\'c}(2005)}]{mezic2005spectral}
\bibinfo{author}{I.~Mezi{\'c}},
\newblock \bibinfo{title}{Spectral properties of dynamical systems, model
  reduction and decompositions},
\newblock \bibinfo{journal}{Nonlinear Dynamics} \bibinfo{volume}{41}
  (\bibinfo{year}{2005}) \bibinfo{pages}{309--325}.
\bibitem[{Budi{\v{s}}i{\'c} et~al.(2012)Budi{\v{s}}i{\'c}, Mohr, and
  Mezi{\'c}}]{budivsic2012applied}
\bibinfo{author}{M.~Budi{\v{s}}i{\'c}}, \bibinfo{author}{R.~Mohr},
  \bibinfo{author}{I.~Mezi{\'c}},
\newblock \bibinfo{title}{Applied koopmanism a},
\newblock \bibinfo{journal}{Chaos: An Interdisciplinary Journal of Nonlinear
  Science} \bibinfo{volume}{22} (\bibinfo{year}{2012}) \bibinfo{pages}{047510}.
\bibitem[{Mezi{\'c}(2013)}]{mezic2013analysis}
\bibinfo{author}{I.~Mezi{\'c}},
\newblock \bibinfo{title}{Analysis of fluid flows via spectral properties of
  the koopman operator},
\newblock \bibinfo{journal}{Annual Review of Fluid Mechanics}
  \bibinfo{volume}{45} (\bibinfo{year}{2013}) \bibinfo{pages}{357--378}.
\bibitem[{Brunton et~al.(2017)Brunton, Brunton, Proctor, Kaiser, and
  Kutz}]{brunton2017chaos}
\bibinfo{author}{S.~L. Brunton}, \bibinfo{author}{B.~W. Brunton},
  \bibinfo{author}{J.~L. Proctor}, \bibinfo{author}{E.~Kaiser},
  \bibinfo{author}{J.~N. Kutz},
\newblock \bibinfo{title}{Chaos as an intermittently forced linear system},
\newblock \bibinfo{journal}{Nature Communications} \bibinfo{volume}{8}
  (\bibinfo{year}{2017}).
\bibitem[{Bongard and Lipson(2007)}]{bongard2007automated}
\bibinfo{author}{J.~Bongard}, \bibinfo{author}{H.~Lipson},
\newblock \bibinfo{title}{Automated reverse engineering of nonlinear dynamical
  systems},
\newblock \bibinfo{journal}{Proceedings of the National Academy of Sciences}
  \bibinfo{volume}{104} (\bibinfo{year}{2007}) \bibinfo{pages}{9943--9948}.
\bibitem[{Schmidt and Lipson(2009)}]{schmidt2009distilling}
\bibinfo{author}{M.~Schmidt}, \bibinfo{author}{H.~Lipson},
\newblock \bibinfo{title}{Distilling free-form natural laws from experimental
  data},
\newblock \bibinfo{journal}{science} \bibinfo{volume}{324}
  (\bibinfo{year}{2009}) \bibinfo{pages}{81--85}.
\bibitem[{Tibshirani(1996)}]{tibshirani1996regression}
\bibinfo{author}{R.~Tibshirani},
\newblock \bibinfo{title}{Regression shrinkage and selection via the lasso},
\newblock \bibinfo{journal}{Journal of the Royal Statistical Society. Series B
  (Methodological)}  (\bibinfo{year}{1996}) \bibinfo{pages}{267--288}.
\bibitem[{Brunton et~al.(2016)Brunton, Proctor, and
  Kutz}]{brunton2016discovering}
\bibinfo{author}{S.~L. Brunton}, \bibinfo{author}{J.~L. Proctor},
  \bibinfo{author}{J.~N. Kutz},
\newblock \bibinfo{title}{Discovering governing equations from data by sparse
  identification of nonlinear dynamical systems},
\newblock \bibinfo{journal}{Proceedings of the National Academy of Sciences}
  \bibinfo{volume}{113} (\bibinfo{year}{2016}) \bibinfo{pages}{3932--3937}.
\bibitem[{Mangan et~al.(2016)Mangan, Brunton, Proctor, and
  Kutz}]{mangan2016inferring}
\bibinfo{author}{N.~M. Mangan}, \bibinfo{author}{S.~L. Brunton},
  \bibinfo{author}{J.~L. Proctor}, \bibinfo{author}{J.~N. Kutz},
\newblock \bibinfo{title}{Inferring biological networks by sparse
  identification of nonlinear dynamics},
\newblock \bibinfo{journal}{IEEE Transactions on Molecular, Biological and
  Multi-Scale Communications} \bibinfo{volume}{2} (\bibinfo{year}{2016})
  \bibinfo{pages}{52--63}.
\bibitem[{Wang et~al.(2011)Wang, Yang, Lai, Kovanis, and
  Grebogi}]{wang2011predicting}
\bibinfo{author}{W.-X. Wang}, \bibinfo{author}{R.~Yang}, \bibinfo{author}{Y.-C.
  Lai}, \bibinfo{author}{V.~Kovanis}, \bibinfo{author}{C.~Grebogi},
\newblock \bibinfo{title}{Predicting catastrophes in nonlinear dynamical
  systems by compressive sensing},
\newblock \bibinfo{journal}{Physical Review Letters} \bibinfo{volume}{106}
  (\bibinfo{year}{2011}) \bibinfo{pages}{154101}.
\bibitem[{Schaeffer et~al.(2013)Schaeffer, Caflisch, Hauck, and
  Osher}]{schaeffer2013sparse}
\bibinfo{author}{H.~Schaeffer}, \bibinfo{author}{R.~Caflisch},
  \bibinfo{author}{C.~D. Hauck}, \bibinfo{author}{S.~Osher},
\newblock \bibinfo{title}{Sparse dynamics for partial differential equations},
\newblock \bibinfo{journal}{Proceedings of the National Academy of Sciences}
  \bibinfo{volume}{110} (\bibinfo{year}{2013}) \bibinfo{pages}{6634--6639}.
\bibitem[{Ozoli{\c{n}}{\v{s}} et~al.(2013)Ozoli{\c{n}}{\v{s}}, Lai, Caflisch,
  and Osher}]{ozolicnvs2013compressed}
\bibinfo{author}{V.~Ozoli{\c{n}}{\v{s}}}, \bibinfo{author}{R.~Lai},
  \bibinfo{author}{R.~Caflisch}, \bibinfo{author}{S.~Osher},
\newblock \bibinfo{title}{Compressed modes for variational problems in
  mathematics and physics},
\newblock \bibinfo{journal}{Proceedings of the National Academy of Sciences}
  \bibinfo{volume}{110} (\bibinfo{year}{2013}) \bibinfo{pages}{18368--18373}.
\bibitem[{Mackey et~al.(2014)Mackey, Schaeffer, and
  Osher}]{mackey2014compressive}
\bibinfo{author}{A.~Mackey}, \bibinfo{author}{H.~Schaeffer},
  \bibinfo{author}{S.~Osher},
\newblock \bibinfo{title}{On the compressive spectral method},
\newblock \bibinfo{journal}{Multiscale Modeling \& Simulation}
  \bibinfo{volume}{12} (\bibinfo{year}{2014}) \bibinfo{pages}{1800--1827}.
\bibitem[{Brunton et~al.(2014)Brunton, Tu, Bright, and
  Kutz}]{brunton2014compressive}
\bibinfo{author}{S.~L. Brunton}, \bibinfo{author}{J.~H. Tu},
  \bibinfo{author}{I.~Bright}, \bibinfo{author}{J.~N. Kutz},
\newblock \bibinfo{title}{Compressive sensing and low-rank libraries for
  classification of bifurcation regimes in nonlinear dynamical systems},
\newblock \bibinfo{journal}{SIAM Journal on Applied Dynamical Systems}
  \bibinfo{volume}{13} (\bibinfo{year}{2014}) \bibinfo{pages}{1716--1732}.
\bibitem[{Proctor et~al.(2014)Proctor, Brunton, Brunton, and
  Kutz}]{proctor2014exploiting}
\bibinfo{author}{J.~L. Proctor}, \bibinfo{author}{S.~L. Brunton},
  \bibinfo{author}{B.~W. Brunton}, \bibinfo{author}{J.~Kutz},
\newblock \bibinfo{title}{Exploiting sparsity and equation-free architectures
  in complex systems},
\newblock \bibinfo{journal}{The European Physical Journal Special Topics}
  \bibinfo{volume}{223} (\bibinfo{year}{2014}) \bibinfo{pages}{2665--2684}.
\bibitem[{Bai et~al.(2014)Bai, Wimalajeewa, Berger, Wang, Glauser, and
  Varshney}]{bai2014low}
\bibinfo{author}{Z.~Bai}, \bibinfo{author}{T.~Wimalajeewa},
  \bibinfo{author}{Z.~Berger}, \bibinfo{author}{G.~Wang},
  \bibinfo{author}{M.~Glauser}, \bibinfo{author}{P.~K. Varshney},
\newblock \bibinfo{title}{Low-dimensional approach for reconstruction of
  airfoil data via compressive sensing},
\newblock \bibinfo{journal}{AIAA Journal}  (\bibinfo{year}{2014}).
\bibitem[{Tran and Ward(2016)}]{tran2016exact}
\bibinfo{author}{G.~Tran}, \bibinfo{author}{R.~Ward},
\newblock \bibinfo{title}{Exact recovery of chaotic systems from highly
  corrupted data},
\newblock \bibinfo{journal}{arXiv preprint arXiv:1607.01067}
  (\bibinfo{year}{2016}).
\bibitem[{Raissi et~al.(2017{\natexlab{a}})Raissi, Perdikaris, and
  Karniadakis}]{raissi2017physics_I}
\bibinfo{author}{M.~Raissi}, \bibinfo{author}{P.~Perdikaris},
  \bibinfo{author}{G.~E. Karniadakis},
\newblock \bibinfo{title}{Physics informed deep learning (part ii): Data-driven
  discovery of nonlinear partial differential equations},
\newblock \bibinfo{journal}{arXiv preprint arXiv:1711.10566}
  (\bibinfo{year}{2017}{\natexlab{a}}).
\bibitem[{Raissi et~al.(2017{\natexlab{b}})Raissi, Perdikaris, and
  Karniadakis}]{raissi2017physics_II}
\bibinfo{author}{M.~Raissi}, \bibinfo{author}{P.~Perdikaris},
  \bibinfo{author}{G.~E. Karniadakis},
\newblock \bibinfo{title}{Physics informed deep learning (part i): Data-driven
  solutions of nonlinear partial differential equations},
\newblock \bibinfo{journal}{arXiv preprint arXiv:1711.10561}
  (\bibinfo{year}{2017}{\natexlab{b}}).
\bibitem[{Basdevant et~al.(1986)Basdevant, Deville, Haldenwang, Lacroix,
  Ouazzani, Peyret, Orlandi, and Patera}]{basdevant1986spectral}
\bibinfo{author}{C.~Basdevant}, \bibinfo{author}{M.~Deville},
  \bibinfo{author}{P.~Haldenwang}, \bibinfo{author}{J.~Lacroix},
  \bibinfo{author}{J.~Ouazzani}, \bibinfo{author}{R.~Peyret},
  \bibinfo{author}{P.~Orlandi}, \bibinfo{author}{A.~Patera},
\newblock \bibinfo{title}{Spectral and finite difference solutions of the
  {B}urgers equation},
\newblock \bibinfo{journal}{Computers \& fluids} \bibinfo{volume}{14}
  (\bibinfo{year}{1986}) \bibinfo{pages}{23--41}.
\bibitem[{Baydin et~al.(2015)Baydin, Pearlmutter, Radul, and
  Siskind}]{baydin2015automatic}
\bibinfo{author}{A.~G. Baydin}, \bibinfo{author}{B.~A. Pearlmutter},
  \bibinfo{author}{A.~A. Radul}, \bibinfo{author}{J.~M. Siskind},
\newblock \bibinfo{title}{Automatic differentiation in machine learning: a
  survey},
\newblock \bibinfo{journal}{arXiv preprint arXiv:1502.05767}
  (\bibinfo{year}{2015}).
\bibitem[{Raissi et~al.(2017)Raissi, Perdikaris, and
  Karniadakis}]{raissi2017machine}
\bibinfo{author}{M.~Raissi}, \bibinfo{author}{P.~Perdikaris},
  \bibinfo{author}{G.~E. Karniadakis},
\newblock \bibinfo{title}{Machine learning of linear differential equations
  using {G}aussian processes},
\newblock \bibinfo{journal}{Journal of Computational Physics}
  \bibinfo{volume}{348} (\bibinfo{year}{2017}) \bibinfo{pages}{683 -- 693}.
\bibitem[{Raissi and Karniadakis(2017)}]{raissi2017hidden}
\bibinfo{author}{M.~Raissi}, \bibinfo{author}{G.~E. Karniadakis},
\newblock \bibinfo{title}{Hidden physics models: Machine learning of nonlinear
  partial differential equations},
\newblock \bibinfo{journal}{arXiv preprint arXiv:1708.00588}
  (\bibinfo{year}{2017}).
\bibitem[{Raissi et~al.(2017{\natexlab{a}})Raissi, Perdikaris, and
  Karniadakis}]{raissi2017inferring}
\bibinfo{author}{M.~Raissi}, \bibinfo{author}{P.~Perdikaris},
  \bibinfo{author}{G.~E. Karniadakis},
\newblock \bibinfo{title}{Inferring solutions of differential equations using
  noisy multi-fidelity data},
\newblock \bibinfo{journal}{Journal of Computational Physics}
  \bibinfo{volume}{335} (\bibinfo{year}{2017}{\natexlab{a}})
  \bibinfo{pages}{736--746}.
\bibitem[{Raissi et~al.(2017{\natexlab{b}})Raissi, Perdikaris, and
  Karniadakis}]{raissi2017numerical}
\bibinfo{author}{M.~Raissi}, \bibinfo{author}{P.~Perdikaris},
  \bibinfo{author}{G.~E. Karniadakis},
\newblock \bibinfo{title}{Numerical {G}aussian processes for time-dependent and
  non-linear partial differential equations},
\newblock \bibinfo{journal}{arXiv preprint arXiv:1703.10230}
  (\bibinfo{year}{2017}{\natexlab{b}}).
\bibitem[{Abadi et~al.(2016)Abadi, Agarwal, Barham, Brevdo, Chen, Citro,
  Corrado, Davis, Dean, Devin et~al.}]{abadi2016tensorflow}
\bibinfo{author}{M.~Abadi}, \bibinfo{author}{A.~Agarwal},
  \bibinfo{author}{P.~Barham}, \bibinfo{author}{E.~Brevdo},
  \bibinfo{author}{Z.~Chen}, \bibinfo{author}{C.~Citro}, \bibinfo{author}{G.~S.
  Corrado}, \bibinfo{author}{A.~Davis}, \bibinfo{author}{J.~Dean},
  \bibinfo{author}{M.~Devin}, et~al.,
\newblock \bibinfo{title}{Tensorflow: Large-scale machine learning on
  heterogeneous distributed systems},
\newblock \bibinfo{journal}{arXiv preprint arXiv:1603.04467}
  (\bibinfo{year}{2016}).
\bibitem[{Driscoll et~al.(2014)Driscoll, Hale, and
  Trefethen}]{driscoll2014chebfun}
\bibinfo{author}{T.~A. Driscoll}, \bibinfo{author}{N.~Hale},
  \bibinfo{author}{L.~N. Trefethen}, \bibinfo{title}{Chebfun guide},
  \bibinfo{year}{2014}.
\bibitem[{Goodfellow et~al.(2016)Goodfellow, Bengio, and
  Courville}]{goodfellow2016deep}
\bibinfo{author}{I.~Goodfellow}, \bibinfo{author}{Y.~Bengio},
  \bibinfo{author}{A.~Courville}, \bibinfo{title}{Deep learning},
  \bibinfo{publisher}{MIT press}, \bibinfo{year}{2016}.
\bibitem[{Kutz et~al.(2016)Kutz, Brunton, Brunton, and
  Proctor}]{kutz2016dynamic}
\bibinfo{author}{J.~N. Kutz}, \bibinfo{author}{S.~L. Brunton},
  \bibinfo{author}{B.~W. Brunton}, \bibinfo{author}{J.~L. Proctor},
  \bibinfo{title}{Dynamic Mode Decomposition: Data-Driven Modeling of Complex
  Systems}, volume \bibinfo{volume}{149}, \bibinfo{publisher}{SIAM},
  \bibinfo{year}{2016}.
\bibitem[{Taira and Colonius(2007)}]{taira2007immersed}
\bibinfo{author}{K.~Taira}, \bibinfo{author}{T.~Colonius},
\newblock \bibinfo{title}{The immersed boundary method: a projection approach},
\newblock \bibinfo{journal}{Journal of Computational Physics}
  \bibinfo{volume}{225} (\bibinfo{year}{2007}) \bibinfo{pages}{2118--2137}.
\bibitem[{Colonius and Taira(2008)}]{colonius2008fast}
\bibinfo{author}{T.~Colonius}, \bibinfo{author}{K.~Taira},
\newblock \bibinfo{title}{A fast immersed boundary method using a nullspace
  approach and multi-domain far-field boundary conditions},
\newblock \bibinfo{journal}{Computer Methods in Applied Mechanics and
  Engineering} \bibinfo{volume}{197} (\bibinfo{year}{2008})
  \bibinfo{pages}{2131--2146}.

\end{thebibliography}







\end{document}